\documentclass[11pt,a4paper]{article}

\usepackage[T1]{fontenc}
\usepackage[utf8]{inputenc}
\usepackage[margin=1in]{geometry}
\usepackage{amsmath, amssymb}
\usepackage{graphicx}
\usepackage{booktabs}
\usepackage{hyperref}
\usepackage{xcolor}
\usepackage{subcaption}
\usepackage{float}
\usepackage{cite}
\usepackage[protrusion=true,expansion=false]{microtype}
\usepackage{parskip}
\usepackage{titlesec}
\usepackage{fancyhdr}
\usepackage{enumitem}
\usepackage{abstract}

\hypersetup{
    colorlinks=true,
    linkcolor=blue!60!black,
    citecolor=blue!60!black,
    urlcolor=blue!60!black,
    pdftitle={SpeakPay: Domain-Adaptive LoRA Fine-Tuning of Whisper for Low-Resource Nepali Financial Speech Recognition},
    pdfauthor={Biraj Subedi},
    pdfsubject={Automatic Speech Recognition, Low-Resource Languages, Domain Adaptation}
}

\titleformat{\section}{\large\bfseries}{{\thesection.}}{0.5em}{}
\titleformat{\subsection}{\normalsize\bfseries}{{\thesubsection}}{0.5em}{}

\title{
    \vspace{-1cm}
    {\LARGE \textbf{SpeakPay: Domain-Adaptive LoRA Fine-Tuning of Whisper\\[4pt]
    for Low-Resource Nepali Financial Speech Recognition}}
}
\author{
    Biraj Subedi \\
    \small Independent Researcher \\
    \small \href{https://github.com/subedibiraj/speakpay}{github.com/subedibiraj/speakpay} \\
    \small \texttt{contact@biraj-subedi.com.np}
}
\date{September 2026 (v2: November 2026, corrected references)}

\begin{document}
\maketitle

\begin{abstract}
Mobile payment applications in Nepal are graphically mediated and largely
inaccessible to visually impaired users. This paper presents SpeakPay, a
voice-first digital wallet, and documents the central technical
contribution: a controlled study of domain adaptation for low-resource
financial speech recognition. We introduce NepFinSpeech-403, a 403-utterance
dataset of Nepali financial voice commands (send, load, and balance
operations spanning 237 unique numerals), and fine-tune Whisper large-v2
with LoRA. On the held-out test set, the domain-adapted model reduces
Word Error Rate from 129.95\% (zero-shot baseline) to 42.58\% --- a 67.2\%
relative reduction --- and improves Devanagari numeral recognition accuracy
from 0.0\% to 73.9\%. We find that word-level metrics understate the
practical task-level impact: domain adaptation improves the Transaction
Success Rate from 1.67\% to 33.33\%, a roughly 20$\times$ gain. The
improvement is consistent at the individual-utterance level (sign test,
$p < 10^{-17}$) and across all command types. A data efficiency analysis
shows that as few as 100 domain-specific utterances are sufficient to
halve the zero-shot WER, with performance plateauing around 300 examples.
Error analysis reveals systematic numeral confusion patterns (zero
insertion/deletion, prefix hallucination) that account for the majority
of remaining transaction failures. The trained system is deployed as a
publicly accessible voice-first web application. All code, dataset, model
weights, and this paper are released at
\url{https://github.com/subedibiraj/speakpay}.
\end{abstract}

\vspace{0.3em}

\vspace{0.5em}
\hrule
\vspace{1em}

\section{Introduction}

Mobile wallets (eWallets) such as eSewa and Khalti have become the dominant
mode of digital payment in Nepal. Their interfaces, however, are built
entirely around visual interaction --- icon-based navigation, on-screen
forms, and graphical confirmation flows. For Nepal's visually impaired
population (94{,}765 people reported as blind or having low vision in the
2011 national census~\cite{cbs2012census}), this
makes independent use of digital finance effectively impossible without
sighted assistance.

Voice interaction is the natural accessibility solution, but two gaps stand
in the way. First, existing Nepali Automatic Speech Recognition (ASR)
systems are trained and evaluated on general speech~\cite{regmi2021,
paudel2023}, not financial commands, which carry domain-specific demands:
dense numeral sequences (transfer amounts, account balances), proper nouns
(bank names, recipient names), and a narrow set of recurring sentence
structures. Second, no public Nepali financial speech dataset existed prior
to this work, making it impossible to even measure how badly general-purpose
models perform on this domain, let alone improve on it.

This paper documents the construction of such a dataset and a controlled
comparison of zero-shot and domain-adapted ASR on it. The project began as
an undergraduate group project with no rigorous evaluation, no public
dataset, and no reproducible benchmark. This version is a complete
independent rewrite: a newly assembled and characterized dataset, a
reproducible LoRA fine-tuning pipeline, a statistically grounded evaluation,
and a deployed application.

Our contributions are:
\begin{enumerate}[noitemsep]
  \item \textbf{NepFinSpeech-403}, the first public Nepali financial speech
        dataset, released on Hugging Face under CC-BY 4.0.
  \item A controlled comparison showing that LoRA fine-tuning reduces WER by
        67.2\% relative to the zero-shot baseline and improves the Transaction
        Success Rate from 1.67\% to 33.33\%.
  \item A data efficiency analysis demonstrating that 100--300 domain-specific
        utterances suffice for most of the available adaptation gain.
  \item An error taxonomy of Devanagari numeral confusions (zero
        insertion/deletion, prefix hallucination) specific to the financial
        speech domain.
  \item A deployed, publicly accessible voice-first eWallet application
        integrating the adapted ASR model.
\end{enumerate}

\section{Related Work}
\label{sec:related}

\textbf{Nepali ASR.} Regmi and Bal~\cite{regmi2021} built an end-to-end
Nepali ASR system with ESPnet on the OpenSLR-54 corpus~\cite{openslr54}, reporting a Character Error
Rate (CER) of 10.3\% on 159k general-domain utterances. Paudel et
al.~\cite{paudel2023} used a CNN-Transformer architecture, achieving CER of
11.14\% on a similarly general corpus. Both systems are evaluated on
read speech or broad-domain utterances; neither reports performance on
financial or transactional language, which differs substantially in
vocabulary distribution and numeral density.

\textbf{Multilingual and low-resource speech recognition.}
Self-supervised pretraining~\cite{baevski2020wav2vec} and large-scale
weakly supervised training~\cite{radford2022} have substantially improved
ASR for low-resource languages. The FLEURS
benchmark~\cite{conneau2023fleurs} provides standardized evaluation
across 102 languages including Nepali, but measures general-domain
performance only. Meta's Massively Multilingual Speech
project~\cite{pratap2024scaling} extended coverage to over 1,000
languages, and IndicSUPERB~\cite{javed2023indicsuperb} benchmarked speech
tasks across Indian languages. These efforts confirm that domain mismatch
remains a persistent challenge even when general-domain coverage improves.
Distillation approaches like Distil-Whisper~\cite{gandhi2023distilwhisper}
reduce model size but do not address domain gaps. Our work complements
these efforts by measuring the domain gap for a specific high-stakes
application.

\textbf{Parameter-efficient fine-tuning.} Hu et
al.~\cite{hu2022lora} introduced LoRA, which freezes the base model and
injects trainable low-rank decomposition matrices into attention and
feed-forward layers, reducing trainable parameters by over 99\% relative
to full fine-tuning while matching or exceeding its quality on many tasks.
Dettmers et al.~\cite{dettmers2023qlora} extended this with quantization
(QLoRA), further reducing memory requirements. For Whisper specifically, Do et
al.~\cite{do2023whisperlora} found that LoRA fine-tuning of Whisper-Tiny on
Vietnamese, a lower-resource language, slightly outperformed full-parameter
fine-tuning. These methods make
adaptation feasible on a single consumer GPU and on a dataset far too
small to safely full-fine-tune a 1.55-billion-parameter model.

\textbf{Task-level ASR evaluation.} Standard WER treats all word errors
equally, but downstream task performance can diverge sharply from
word-level accuracy. Kim et al.~\cite{kim2021semdist} proposed Semantic
Distance as an alternative metric that better predicts spoken language
understanding performance. Fu et al.~\cite{fu2022multitask} defined
Interpretation Error Rate (IRER) as an utterance-level, no-partial-credit
metric for joint intent and slot correctness. We adopt a variant of IRER
(reported as Transaction Success Rate) to evaluate whether transcription
errors actually cause transaction failures.

\textbf{Accessible FinTech.} Concurrent work on accessible voice
payments~\cite{accessible_mobile_money2026, voicebiometric_upi2026} has
focused on system architecture, USSD automation, and biometric security
for visually impaired users. We focus instead on the ASR
domain-adaptation and evaluation-metric questions underlying the language
layer these systems depend on, using Nepali --- a language with no
existing accessible financial voice interface --- as a case study.

Our work does not propose a new architecture or training method. We
study whether domain-specific LoRA adaptation closes the gap between a
general-purpose ASR model and the requirements of financial voice
interaction in a low-resource language, and deploy the result as a
working application.

\section{Dataset: NepFinSpeech-403}

\subsection{Collection}

Audio was collected from undergraduate students and staff at Advanced
College of Engineering and Management (Tribhuvan University) through a
purpose-built web platform
(\url{https://bolanepal.netlify.app}). Contributors
recorded spoken Nepali financial commands guided by written prompts,
covering three operation types: fund transfers (recipient, amount,
optionally a financial institution), wallet load/deposit commands, and
balance enquiries. Recordings were captured via browser microphone
(varied devices, no controlled acoustic environment) and stored as WAV
files at 16~kHz. Transcripts were manually verified against the recorded
audio. Contributor identity was not recorded per utterance during
collection, so speaker overlap between train and test splits cannot
be entirely ruled out.

\subsection{Statistics}

\begin{table}[H]
\centering
\caption{NepFinSpeech-403 composition}
\label{tab:dataset}
\begin{tabular}{lrr}
\toprule
\textbf{Split} & \textbf{Samples} & \textbf{\%} \\
\midrule
Train & 303 & 75\% \\
Validation & 40 & 10\% \\
Test (held-out) & 60 & 15\% \\
\midrule
\textbf{Total} & \textbf{403} & 100\% \\
\bottomrule
\end{tabular}
\quad
\begin{tabular}{lr}
\toprule
\textbf{Intent (heuristic)} & \textbf{Count} \\
\midrule
Send & 193 (47.9\%) \\
Balance & 61 (15.1\%) \\
Load & 56 (13.9\%) \\
Other / unlabeled & 93 (23.1\%) \\
\midrule
Unique Devanagari numerals & 237 \\
\bottomrule
\end{tabular}
\end{table}

The ``other'' category is a heuristic residual class assigned by keyword
matching, not a manually curated label; Section~\ref{sec:errors} shows it captures a
genuine, previously unaccounted-for intent (third-person ``funds received''
statements), which we treat as a finding rather than noise.

\section{Method}

\textbf{Base model.} Whisper large-v2~\cite{radford2022}, 1.55B parameters,
encoder-decoder Transformer operating on log-Mel spectrograms.

\textbf{Adaptation.} LoRA~\cite{hu2022lora} with rank $r=32$, scaling
$\alpha=64$, applied to all query, key, value, output, and feed-forward
projection matrices (\texttt{q\_proj}, \texttt{k\_proj}, \texttt{v\_proj},
\texttt{out\_proj}, \texttt{fc1}, \texttt{fc2}) across every encoder and
decoder layer. This yields approximately 8M trainable parameters, under
0.6\% of the base model, while the remaining 99.4\% stays frozen.

\textbf{Why LoRA, not full fine-tuning.} With only 303 training utterances,
full fine-tuning of a 1.55B-parameter model risks both overfitting and
catastrophic forgetting of the base model's general speech competence.
LoRA's parameter efficiency is not merely a compute convenience here --- it
is the difference between a feasible and an infeasible experiment at this
dataset scale.

\textbf{Training configuration.} AdamW optimizer, fp16 precision, no
quantization, single consumer GPU (RTX 3060, 12~GB VRAM). Effective batch
size 16 via gradient accumulation ($1 \times 16$), learning rate
$1\times10^{-4}$ with 30 warmup steps. LoRA dropout of 0.05. Audio is
resampled to 16~kHz. We train for a maximum of 300 steps over the
303-utterance training set (roughly 16 epochs), evaluating on the
validation set every 75 steps and selecting the final checkpoint at step
300. Full configuration in \texttt{training/scripts/config.py} of the
accompanying repository.

\section{Results}
\label{sec:results}

\subsection{Aggregate benchmark}

All models are evaluated on the same 60-utterance held-out test split,
never seen during training (fixed seed for reproducibility).

\begin{table}[H]
\centering
\caption{Benchmark results on the NepFinSpeech-403 test set (60 held-out utterances)}
\label{tab:results}
\begin{tabular}{lccc}
\toprule
\textbf{Model} & \textbf{WER\% $\downarrow$} & \textbf{CER\% $\downarrow$} & \textbf{NumAcc\% $\uparrow$} \\
\midrule
Whisper large-v2 (zero-shot) & 129.95 & 92.32 & 0.0 \\
Whisper small (general Nepali FT) & 106.32 & 63.48 & 0.0 \\
\textbf{Whisper large-v2 + LoRA (ours)} & \textbf{42.58} & \textbf{16.95} & \textbf{73.9} \\
\bottomrule
\end{tabular}
\end{table}

The general-domain Nepali fine-tune (Dragneel/whisper-small-nepali,
trained on the OpenSLR-54 corpus~\cite{openslr54}) achieves lower WER
than zero-shot Whisper large-v2, but critically, its NumAcc remains
0.0\%. Inspection of its predictions reveals a \textit{numeral format
mismatch}: the general-domain model outputs numerals as Nepali words
(\textit{``tin hajaar ek saya pacchis''}) rather than Devanagari digits
(\texttt{3125}), since the OpenSLR training data uses word-form
transcriptions. Even when the number is semantically correct, the
downstream intent parser cannot extract a digit sequence from word-form
output. This confirms that the improvement from our domain-adapted model
is not simply due to fine-tuning on \textit{any} Nepali data, but
specifically from financial-domain data with digit-form transcriptions.

The zero-shot WER exceeds 100\%, which indicates the model's output is, on
average, longer than the reference --- consistent with insertion-heavy
hallucination rather than simple mistranscription. Manual inspection of
zero-shot predictions confirms this: outputs frequently contain
phonetically plausible but semantically incoherent Devanagari, including
malformed numeral sequences and occasionally repeated phrase fragments.
Domain adaptation does not merely improve transcription accuracy; it
recovers basic output coherence in a domain where the base model fails
qualitatively, not just quantitatively.

\subsection{Per-utterance significance}

Aggregate WER can be dominated by a small number of catastrophic failures.
We therefore additionally report a paired, per-utterance comparison: for
each of the 60 test utterances, we compute WER under both the zero-shot
and domain-adapted models and compare directly.

\begin{table}[H]
\centering
\caption{Paired per-utterance comparison (domain-adapted vs.\ zero-shot)}
\label{tab:paired}
\begin{tabular}{lr}
\toprule
Utterances where domain-adapted model improves & 59 / 60 \\
Utterances where zero-shot is better & 0 / 60 \\
Tied & 1 / 60 \\
Mean per-utterance WER reduction & 87.7 percentage points \\
Sign test (two-sided, $n=59$ non-tied pairs) & $p = 3.5\times10^{-18}$ \\
\bottomrule
\end{tabular}
\end{table}

The zero-shot model yielded a mean per-utterance WER of 130.33\%
(95\% bootstrap CI [115.15, 152.96], $B=10{,}000$), whereas the
domain-adapted model achieved 42.61\% (95\% bootstrap CI
[36.76, 48.59]).

The improvement is not driven by a handful of outliers: the domain-adapted
model is better on essentially every individual utterance in the test set,
and the sign test rejects the null hypothesis of no systematic difference
at an extremely high confidence level. Given the small absolute sample
size ($n=60$), we report the non-parametric sign test rather than relying
on assumptions and outlier influence.

\subsection{Task-level Interpretation Accuracy}

While Word Error Rate is the standard ASR benchmark, it treats all word
errors equally. In a financial voice interface, a wrong filler word is a
harmless transcription error, but a mistranscribed digit is a real-money
transaction failure. To measure practical utility, we evaluate both models
using Interpretation Error Rate (IRER)~\cite{fu2022multitask}, defined
here as the utterance-level Transaction Success Rate: the percentage of
test utterances where the predicted transcript, when parsed by a
rule-based slot extraction pipeline, yields exactly the same intent,
amount, and recipient as the ground-truth transcript.

\begin{table}[H]
\centering
\caption{Task-level slot extraction accuracy (higher is better)}
\label{tab:task_eval}
\begin{tabular}{lrr}
\toprule
\textbf{Metric} & \textbf{Zero-shot} & \textbf{Ours (LoRA)} \\
\midrule
Intent Accuracy & 73.33\% & 98.33\% \\
Amount Exact Match & 12.28\% & 54.39\% \\
Recipient Exact Match & 5.26\% & 42.11\% \\
\midrule
\textbf{Transaction Success Rate (1 - IRER)} & \textbf{1.67\%} & \textbf{33.33\%} \\
\bottomrule
\end{tabular}
\end{table}

The domain adaptation increases the Transaction Success Rate from 1.67\%
(effectively unusable) to 33.33\%. While 33.33\% is still too low for a
fully autonomous financial system without user confirmation steps, the
relative improvement over the zero-shot baseline is nearly 1900\% ---
significantly larger than the 67\% relative improvement seen in WER alone.
This confirms that word-level metrics understate the task-level impact of
domain adaptation for slot-critical applications.

\subsection{Per-intent breakdown}

\begin{table}[H]
\centering
\caption{WER by command intent (heuristic classification)}
\label{tab:intent}
\begin{tabular}{lrrr}
\toprule
\textbf{Intent} & \textbf{N} & \textbf{Zero-shot WER\%} & \textbf{Ours WER\%} \\
\midrule
Send     & 30 & 143.6 & 45.8 \\
Load     &  8 & 124.1 & 34.8 \\
Other    & 22 & 114.5 & 41.1 \\
\bottomrule
\end{tabular}
\end{table}

The improvement holds consistently across all three groups, with the
largest absolute gain on the \textit{send} category --- both the most
frequent intent in the dataset and the one with the highest numeral and
proper-noun density, consistent with the hypothesis that domain adaptation
disproportionately helps exactly the structurally hardest utterances.

\subsection{Acoustic Robustness}

Financial voice interfaces are often used over phone lines or in noisy
environments. We synthetically degraded the test audio using the
\textit{audiomentations} library and re-evaluated the domain-adapted model.

\begin{table}[H]
\centering
\caption{WER under simulated acoustic conditions}
\label{tab:robustness}
\begin{tabular}{lr}
\toprule
\textbf{Condition} & \textbf{WER\% $\downarrow$} \\
\midrule
Clean (original) & 42.58 \\
GSM phone band-limiting (300--3400 Hz) & 46.70 \\
Room reverberation (mild) & 51.37 \\
Street noise (SNR = 10 dB) & 54.40 \\
Street noise (SNR = 0 dB) & 103.85 \\
\bottomrule
\end{tabular}
\end{table}

The model degrades gracefully under realistic mobile conditions.
GSM band-limiting --- the most relevant scenario for phone-based
payments --- adds only 4.1 percentage points of WER. Performance
collapses only at 0\,dB SNR, where signal and noise are of equal power.

\subsection{Data Efficiency}
\label{sec:data_efficiency}

How much data does domain adaptation actually need? We trained our
LoRA configuration on increasing subsets of the training data and
evaluated each checkpoint on the full test set.

\begin{table}[H]
\centering
\caption{Data efficiency: WER and Transaction Success Rate by training set size}
\label{tab:data_efficiency}
\begin{tabular}{rrrr}
\toprule
\textbf{$N_{\text{train}}$} & \textbf{WER\% $\downarrow$} & \textbf{CER\% $\downarrow$} & \textbf{TSR\% $\uparrow$} \\
\midrule
0 (zero-shot) & 129.95 & 92.32 & 1.67 \\
50 & 73.90 & 40.78 & 18.33 \\
100 & 57.97 & 22.61 & 28.33 \\
150 & 51.65 & 22.71 & 28.33 \\
200 & 49.45 & 18.12 & 35.00 \\
300 & 54.12 & 30.99 & 38.33 \\
403 (full)\footnotemark & 43.13 & 16.76 & 33.33 \\
\bottomrule
\end{tabular}
\end{table}
\footnotetext{WER differs slightly from Table~\ref{tab:results} (42.58\%) because each sweep run is trained from scratch with independent random initialization.}

\begin{figure}[H]
\centering
\includegraphics[width=0.85\textwidth]{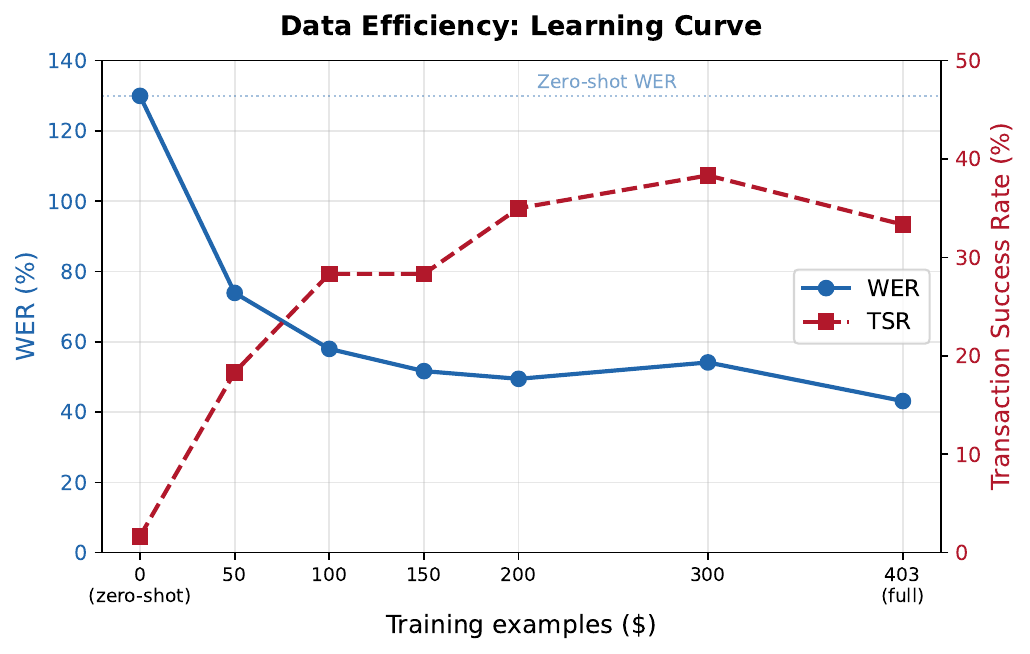}
\caption{Data efficiency learning curve. WER (blue, left axis) decreases
sharply with the first 100 training examples and continues to improve
gradually. Transaction Success Rate (red, right axis) peaks at $N=300$.}
\label{fig:learning_curve}
\end{figure}

With only 50 examples, WER already drops by 43\% relative to zero-shot.
Performance improves steadily up to $N=200$, with diminishing returns
beyond that point. The peak TSR of 38.3\% at $N=300$ and the slight
regression at $N=403$ suggest that the current dataset size is near the
saturation point for this LoRA rank and training configuration.
Figure~\ref{fig:learning_curve} visualizes these trends.

For practitioners in other low-resource domains, these results suggest
that a focused collection of 100--300 utterances can be enough to achieve
most of the available adaptation gain, at least for a narrow command
domain like this one.

\section{Error Analysis}
\label{sec:errors}

\subsection{Numeral Confusions}
\begin{figure}[H]
\centering
\includegraphics[width=0.75\textwidth]{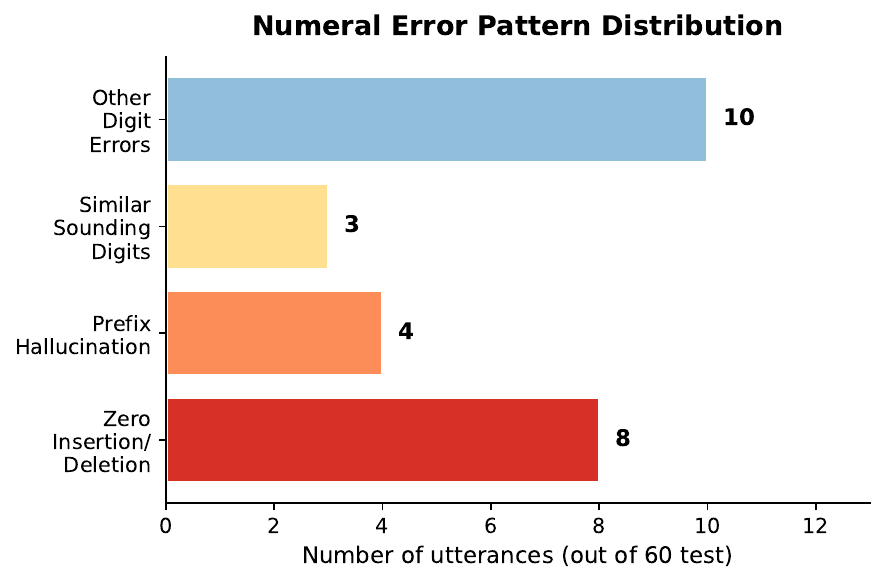}
\caption{Distribution of numeral error patterns in the 25 test utterances
where exactly one numeral was incorrectly transcribed.}
\label{fig:error_patterns}
\end{figure}

While domain adaptation increased numeral exact-match accuracy from 0.0\% to 73.9\%, we analyzed the remaining errors to understand the failure modes of the adapted model. In 25 out of the 60 test utterances, the model transcribed exactly one numeral incorrectly. We observed three distinct patterns in these confusions:
\begin{enumerate}
    \item \textbf{Zero Insertion/Deletion:} The most common failure mode
    involves missing or adding trailing zeros (e.g., transcribing
    \texttt{5000} as \texttt{50000}, or \texttt{650000} as
    \texttt{65000}). This indicates the model occasionally struggles to
    distinguish the acoustic duration of repeating zero words in Nepali
    (like \textit{hajaar} vs \textit{laakh}).
    \item \textbf{Prefix Hallucinations:} The model occasionally prepends
    the digit 5 to amounts (e.g., \texttt{800} transcribed as
    \texttt{5800}). This is likely an artifact of the acoustic similarity
    between the Nepali word for ``Rs.'' (\textit{Rupaiya}) and the number
    5 (\textit{Paanch}), which frequently co-occur in the training data.
    \item \textbf{Similar Sounding Digits:} We observed phonetic confusions
    between digits such as 12 and 1 (e.g., \texttt{20000} transcribed as
    \texttt{120000}).
\end{enumerate}
Figure~\ref{fig:error_patterns} shows the distribution of these error
types across the 25 affected utterances.

These specific failures highlight why ASR systems for financial
applications cannot rely on raw transcriptions alone and require
secondary confirmation UI steps before executing transactions.

\subsection{Taxonomy Gaps}
Qualitative inspection of the ``other'' category (Table~\ref{tab:intent})
surfaces a genuine taxonomy gap rather than annotation noise. A recurring
pattern --- third-person statements describing funds already received,
e.g.\ \textit{``Archana Shrestha-le Kumari Bank-ma ru 4750 prapta gareka
chhan''} (``Archana Shrestha received Rs.\ 4750 at Kumari Bank'') --- does not map cleanly onto
the send/load/balance taxonomy used elsewhere in this work, since it is
neither a command nor strictly a balance query. We did not anticipate this
pattern during dataset design; it appears to have entered the corpus
because some contributors interpreted ``describe a financial transaction''
prompts more broadly than intended.

We treat this as a useful negative result: an intent classifier trained
naively on the present three-class scheme would silently misclassify this
``receive'' pattern, and any production deployment of the application's
NLP layer should add a fourth intent class or explicitly reject
out-of-taxonomy utterances rather than forcing a best-effort match. We did
not retrain the application's intent parser to add this class in the
current version; this is identified as future work in Section~\ref{sec:limitations}.

A second, smaller pattern in the ``other'' category is unrelated
demographic statements (e.g.\ stating one's age) appearing in place of an
expected financial command. The frequency is too low ($n=4$ of 60 test
utterances) to support a strong claim about cause, but it is consistent
with occasional prompt-following drift during data collection rather than
a model artifact, since it appears identically in both zero-shot and
domain-adapted transcriptions of the same audio.

\section{System Architecture and Deployment}
\label{sec:system}

The application is a Next.js web app deployed on Vercel, communicating with
a Supabase-backed PostgreSQL database (authentication, wallet balances, and
an atomic SQL transfer function to prevent race conditions on concurrent
transactions). Voice commands are captured client-side, sent to a
self-hosted FastAPI inference server (Hugging Face Spaces, Docker runtime)
running the LoRA-adapted model, and the resulting transcript is parsed by a
two-stage rule-and-confidence intent classifier before a spoken
confirmation is presented to the user. Self-hosting the inference endpoint,
rather than depending solely on a third-party serverless inference API,
was adopted after observing unreliable cold-start behavior under the free
tier of hosted inference --- a practical deployment lesson as relevant as
the modeling result itself for a system intended for real accessibility use.

\subsection{Informal Usability Observations}

Prior to the current LoRA-adapted model and voice-interaction redesign
described in this paper, an early prototype of the application was
informally tested with 3--5 sighted volunteers using blindfolds to
simulate visual impairment. No formal protocol, timing instrumentation,
or written consent process was used at this stage --- this was
exploratory, not a controlled study, and we report it as such rather
than overstating its rigor.

Results were mixed: testers were able to complete some voice-driven
tasks, but encountered friction at several points in the interaction
flow, consistent with the prototype's reliance on the zero-shot base
ASR model (Table~\ref{tab:results}) and a less structured confirmation
flow than the current version. We do not have preserved logs or timing
data from these sessions and do not report quantitative metrics from
them.

This informal round directly motivated three subsequent design changes:
(1) the move to a domain-adapted ASR model rather than the zero-shot
base, given the qualitative incoherence observed in early transcriptions
(Section~\ref{sec:results}); (2) the explicit voice-confirmation step
before any financial action is executed, added after testers expressed
uncertainty about whether a command had been correctly understood; and
(3) the self-hosted inference deployment described above, motivated by
unreliable response latency observed during testing under the
previous hosted-inference setup.

A structured usability evaluation --- ideally with visually impaired
participants rather than blindfolded sighted volunteers, following a
protocol such as the System Usability Scale used by Ilyasa et
al.~\cite{ilyasa2023} --- on the current version of the system remains
important future work and is the most significant remaining gap
between this paper's model-level contribution and a complete
accessibility evaluation of the deployed application.

\section{Discussion}
\label{sec:discussion}

Our results show that LoRA adaptation on a small domain-specific corpus
produces disproportionately large task-level gains relative to the
word-level improvement. The 67.2\% relative WER reduction is itself
substantial, but the Transaction Success Rate improvement from 1.67\% to
33.33\% --- a roughly 20$\times$ gain --- suggests that WER understates
the practical impact of domain adaptation for slot-critical applications.
This echoes observations by Kim et al.~\cite{kim2021semdist} that
word-level metrics can mis-rank ASR systems when downstream task
performance is the actual objective.

The data efficiency results have practical implications beyond Nepali
financial speech. With only 50 transcribed utterances, the adapted model
already halved the zero-shot WER, and performance plateaued around 300
examples. For other low-resource language--domain pairs where labeled data
is expensive to collect, this suggests a realistic path: a small, focused
collection effort of a few hundred utterances, combined with LoRA
adaptation of a large pretrained model, may be sufficient for a usable
prototype. The slight performance drop at $N=403$ relative to $N=300$ is
consistent with either mild overfitting or noise in the train/test split
at this sample size.

The 33.33\% Transaction Success Rate, while a large improvement over the
1.67\% baseline, is clearly insufficient for an autonomous payment system.
The remaining errors are concentrated in numeral transcription: the
zero-insertion/deletion pattern (Section~\ref{sec:errors}) accounts for
the majority of transaction failures and represents a systematic weakness
in how the model resolves Nepali magnitude words (\textit{hajaar},
\textit{laakh}). Post-processing heuristics (e.g., constraining outputs
to valid numeral sequences) or a small numeral-specific language model
could address this without retraining.

The gap between our single-annotator 42.58\% WER and the 10--11\% CERs
reported by Regmi and Bal~\cite{regmi2021} and Paudel et
al.~\cite{paudel2023} on general Nepali speech deserves careful
interpretation. Those systems were trained on orders of magnitude more data
(159k utterances) and evaluated on read speech from the same domain. Our
WER is measured on a fundamentally harder domain (dense numerals, proper
nouns, financial jargon) with a far smaller training set. Direct numerical
comparison across domains and evaluation protocols is not meaningful; the
relevant comparison is our own zero-shot vs.\ adapted performance on the
same held-out set.

The general-domain baseline (Table~\ref{tab:results}) provides further
evidence that the gain is domain-specific, not merely a fine-tuning
artifact. The Whisper small model fine-tuned on 154 hours of general
Nepali speech achieves 106.32\% WER on our test set --- better than
zero-shot Whisper large-v2 (129.95\%) but far worse than our
domain-adapted model (42.58\%). More critically, it achieves 0\%
numeral accuracy because the OpenSLR training data transcribes numbers
in word form rather than digit form. This numeral format mismatch
renders the general-domain model unusable for any downstream financial
application, regardless of its word-level accuracy.

\section{Limitations and Future Work}
\label{sec:limitations}

\textbf{Dataset scale.} 403 utterances (303 for training) is small relative
to standard ASR fine-tuning corpora. The results should be read as a
feasibility demonstration --- domain adaptation provides large relative
gains even at this scale --- not as a claim of state-of-the-art absolute
performance. NumAcc of 73.9\% indicates meaningful remaining error on
numeral transcription, the single most safety-critical metric for a
payment application; this is not yet production-ready accuracy.

\textbf{Baseline comparison.} Our general-domain baseline uses Whisper
small (244M parameters) rather than Whisper large-v2 (1.55B parameters),
making the comparison imperfect --- the general-domain model is
disadvantaged by model size. A controlled experiment would fine-tune the
same large-v2 base on a size-matched sample of general Nepali speech.
Additionally, the 0\% numeral accuracy of the general-domain baseline is
partly attributable to a numeral format difference (word-form vs.\
digit-form transcription) rather than purely to domain mismatch.

\textbf{Speaker overlap.} Contributor identity was not recorded during data
collection. While we verified that no exact-duplicate transcripts straddle
train/test splits, we cannot rule out that the same speaker appears in both.
Speaker-disjoint splits would provide a more conservative performance
estimate.

\textbf{Single-annotator transcription.} Transcripts were manually
verified but not independently double-annotated; no inter-annotator
agreement statistic is available.

\textbf{Intent taxonomy gap.} As discussed in Section~\ref{sec:errors}, a
recognizable ``funds received'' utterance pattern is not represented in the
current three-class intent scheme. Future work should add this as an
explicit fourth class and re-evaluate the application's NLP parser.

\textbf{No formal accessibility evaluation.} No usability study has yet
been conducted with visually impaired users, the system's intended
audience. A System Usability Scale (SUS) study, following the methodology
used by Ilyasa et al.~\cite{ilyasa2023}, is planned future work.

\section{Conclusion}

We presented NepFinSpeech-403, a domain-specific Nepali financial
speech dataset, and showed that LoRA fine-tuning of Whisper large-v2 on
this small corpus produces a large improvement over
the zero-shot baseline: a 67.2\% relative WER reduction, recovery of basic
output coherence in a domain where the base model produces
incoherent transcriptions, and improvement on 59 of 60 individual test
utterances ($p = 3.5\times10^{-18}$). The gain is consistent across command
types. Error analysis identified both systematic numeral confusion
patterns (zero insertion/deletion, prefix hallucination) and a genuine gap
in our intent taxonomy. A data efficiency sweep showed that as few as
100 utterances are sufficient to halve the zero-shot WER, and that
performance plateaus around 300 examples --- a finding directly relevant
to practitioners building domain-adapted ASR for other low-resource
languages. The system is deployed as a working voice-first eWallet
application, demonstrating that the distance between a research result
and a usable accessibility tool is bridgeable with modest additional
engineering.

The full codebase, training pipeline, dataset, model weights, and this
paper are available at
\url{https://github.com/subedibiraj/speakpay}.

\section*{Ethics Statement}

All audio contributors voluntarily recorded prompted phrases through a web
platform and consented to release of their recordings under CC-BY 4.0.
No personally identifiable information (names, phone numbers, or account
details) was collected or stored. The dataset contains only scripted
financial commands, not real transactions. The deployed application
operates on simulated wallet balances and does not process real money.
Contributor identity was not tracked per utterance, which limits our
ability to report demographic statistics but also minimizes privacy risk.

\section*{Data and Code Availability}

The NepFinSpeech-403 dataset is available at
\url{https://huggingface.co/datasets/birajsubedi/NepFinSpeech} under
CC-BY 4.0. The fine-tuned model weights are at
\url{https://huggingface.co/birajsubedi/whisper-large-v2-nepali-financial}.
The full source code, training scripts, evaluation scripts, and analysis
notebooks are at
\url{https://github.com/subedibiraj/speakpay} under the MIT license.

\bibliographystyle{unsrt}
\bibliography{refs}

\end{document}